\documentclass[11pt,a4paper]{article}

\usepackage{times,latexsym}
\usepackage{url}
\usepackage[T1]{fontenc}

\usepackage[acceptedWithA]{tacl2021v1}

\usepackage{amsmath}
\usepackage{amssymb}
\usepackage{booktabs}
\usepackage{graphicx}
\usepackage{multirow}
\usepackage{makecell}
\usepackage{enumitem}
\usepackage{bm}
\usepackage{tabularx}
\usepackage{array}

\usepackage[most]{tcolorbox}

\newtcolorbox{rubricbox}[1]{
  enhanced,
  width=\textwidth,
  colback=gray!6,
  colframe=black,
  colbacktitle=black,
  coltitle=white,
  title={#1},
  fonttitle=\bfseries,
  fontupper=\small,
  boxrule=0.7pt,
  arc=1.5mm,
  left=2.5mm,
  right=2.5mm,
  top=2mm,
  bottom=2mm,
  before skip=0pt,
  after skip=0pt
}

\title{CSPF: A Constrained Shared-Private Fusion Method for Non-Verifiable Preference Evaluation}

\author{
Hehao Zhang \quad
Danli Wang\thanks{\hspace{0.6em}Corresponding author.} \quad
Xinyuan Wang \quad
Xuange Gao
\\
Institute of Automation, Chinese Academy of Sciences,
Beijing 100190, China
\\
School of Artificial Intelligence,
University of Chinese Academy of Sciences,
Beijing 100049, China
\\
\texttt{zhanghehao2023@ia.ac.cn} \quad
\texttt{danli.wang@ia.ac.cn}
\\
\texttt{wangxinyuan2024@ia.ac.cn} \quad
\texttt{gaoxuange2022@ia.ac.cn}
}

\date{}

\begin{document}
\maketitle

\begin{abstract}
At present, reliable evaluation of non-verifiable tasks remains challenging. Existing approaches often fail to adequately capture the diverse evaluative criteria underlying human preferences in such tasks.
To this end, we propose Constrained Shared-Private Fusion (\textsc{CSPF}), a fusion method that treats heterogeneous frozen reward models as complementary evaluators and learns to integrate their hidden-state representations under pairwise human-preference supervision.
\textsc{CSPF} decomposes each expert signal into shared and expert-private representations, encouraging cross-expert alignment while preserving complementary viewpoints.
Across experiments on LM-Arena target-domain adaptation and PPE out-of-distribution preference evaluation, \textsc{CSPF} achieves the best performance on the primary metrics among the evaluated single-expert reward-model, scalar-score multi-expert, and rubric-judge baselines.
Overall, \textsc{CSPF} suggests that fusing hidden-state representations provides a more expressive basis for preference assessment, offering a practical route toward integrated evaluative signals for non-verifiable preference tasks.
\end{abstract}

\section{Introduction}

After pretraining, LLMs are often further adapted through post-training to improve target capabilities.
Reinforcement learning is an important post-training approach: it optimizes a policy model by using evaluators, such as reward models, verifiers, or checkers, to assess generated responses and guide optimization~\citep{ouyangTrainingLanguageModels2022}.
This dependence on assessment quality makes the source of evaluation crucial.
In verifiable domains such as mathematical reasoning and code generation, objective ground-truth answers or executable tests make assessment relatively straightforward, enabling reinforcement learning with rule-based or test-based rewards to improve mathematical reasoning and coding abilities~\citep{shaoDeepSeekMathPushingLimits2024,gehringRLEFGroundingCode2024}.
By contrast, many alignment-relevant tasks are non-verifiable: open-ended dialogue, creative writing, subjective question answering, and safety-sensitive instruction following often lack a single ground-truth standard, and their quality depends on multi-criteria human judgments~\citep{jiaWritingZeroBridgeGap2025,gunjalRubricsRewardsReinforcement2025}.
Reliable evaluation for such tasks is therefore difficult but essential, because post-training can only optimize LLM behavior toward objectives that the evaluator can accurately assess.

Existing work has developed three main evaluator families for non-verifiable preference tasks: preference-based reward models, rubric-based evaluators, and multi-evaluator methods.
Preference-based reward models learn holistic reward functions from human comparisons, rankings, or ratings, providing scalable supervision for LLM post-training~\citep{malikRewardBench2Advancing2025}.
Rubric-based evaluators, often implemented as LLM judges, make criteria explicit and score or compare responses under natural-language rubrics~\citep{gunjalRubricsRewardsReinforcement2025}.
Multi-evaluator methods select, route among, or aggregate heterogeneous reward models to exploit complementary strengths across evaluators~\citep{wuRewardModelRouting2025,wangTransformingCombiningRewards2024}.
These approaches move beyond direct correctness checking, but they differ in how evaluative information is represented: holistic preference labels, explicit natural-language criteria, or final scalar scores from multiple experts.

Despite this progress, existing mechanisms remain limited when preferences are composite. In preference-based reward modeling, pairwise comparisons and scalar rewards provide scalable supervision, but they collapse multidimensional quality judgments into holistic labels. Such holistic signals offer limited fine-grained credit assignment, leaving unclear which criteria drive the preference, which spans are problematic, or where revision is needed \citep{wuFineGrainedHumanFeedback2023}. Rubric-based evaluators make criteria explicit, but natural-language rubrics can be incomplete, overlapping, or misaligned with the intended preference direction. Their scores still depend on how the judge model interprets each criterion, which may introduce misalignment with human preferences \citep{shenRethinkingRubricGeneration2026}. Methods that use multiple evaluators address heterogeneity, but selection and routing reduce feedback to a chosen expert and may miss complementary criteria that should be considered jointly. Scalar aggregation uses multiple experts, yet operates on final scores whose scales and semantics are shaped by different data sources, objectives, and calibration regimes \citep{nguyenLASeRLearningAdaptively2026,wuRewardModelRouting2025,wangTransformingCombiningRewards2024}. These limitations motivate evaluation mechanisms that integrate multidimensional evidence in a semantically aligned representation space and model interactions across criteria, rather than relying solely on holistic labels, judge-dependent rubric scores, or uncalibrated scalar rewards.

To this end, we introduce Constrained Shared-Private Fusion (\textsc{CSPF}), a hidden-state fusion method for evaluating LLM responses in non-verifiable preference tasks.
\textsc{CSPF} treats frozen reward models as complementary evaluative perspectives and learns, under pairwise human-preference supervision, to fuse their hidden-state representations rather than relying only on final scalar scores.
Its constrained shared-private fusion structure separates cross-expert shared representations from expert-private representations, encouraging common preference-relevant signals to align while preserving complementary expert viewpoints.
Because all reward-expert backbones remain frozen, adaptation is localized within the fusion module, making the approach modular and extensible to newly released or domain-specialized reward models.

Our work makes three main contributions:
\begin{enumerate}[label=(\arabic*), leftmargin=*, itemsep=0.25em, topsep=0.2em]
    \item \textbf{\textsc{CSPF} as a method for non-verifiable preference evaluation.}
    We propose \textsc{CSPF}, a hidden-representation-level method for fusing multiple reward models, which implicitly models interactions among latent evaluative factors to improve non-verifiable preference evaluation.

    \item \textbf{Hidden-state fusion mechanism beyond scalar scores.}
    We compare hidden-state and scalar-score fusion methods, showing that representation-level fusion achieves stronger performance and supports more sample-dependent use of complementary reward experts.

    \item \textbf{Experimental validation of fusion design factors.}
    We analyze signal representation, expert-pool composition, and fusion structure, showing that effective multi-expert evaluation depends on coordinated design choices rather than simply adding more experts or scores.
\end{enumerate}

\section{Related Work}
\label{sec:related-work}

\subsection{Preference-Based Reward Models}
\label{sec:rw-reward-modeling}

Preference-based reward models convert human comparisons, rankings, or ratings into learned scalar evaluators for LLM post-training~\citep{ouyangTrainingLanguageModels2022}.
Recent open models illustrate this ecosystem: Skywork-Reward-V2 emphasizes scalable preference-data curation and general-purpose reward modeling~\citep{liuSkyworkRewardV2ScalingPreference2025}, OffsetBias improves robustness against evaluation bias~\citep{parkOffsetBiasLeveragingDebiased2024}, and ArmoRM combines multiple interpretable reward objectives before producing an overall preference score~\citep{wangInterpretablePreferencesMultiObjective2024a}.
Benchmarks such as RewardBench, RewardBench 2, and PPE assess reward models across instruction following, reasoning, safety, and human-preference settings~\citep{lambertRewardBenchEvaluatingReward2025,malikRewardBench2Advancing2025,frickHowEvaluateReward2025}.
Despite this progress, holistic preference RMs collapse multiple quality dimensions into a single score, obscuring the criteria and trade-offs underlying each judgment. Consequently, improvements in aggregate preference can mask regressions in specific dimensions \citep{wuFineGrainedHumanFeedback2023}.

\subsection{Rubric-Based LLM Evaluators}
\label{sec:rw-rubric-evaluators}

Rubric-based LLM evaluators make criteria explicit by prompting or training judge models to score, compare, or critique responses under natural-language rubrics.
Representative open evaluators include Prometheus-2, which supports direct assessment and pairwise ranking with user-defined criteria~\citep{kimPrometheus2Open2024}, and R3, which develops rubric-agnostic reward models with reasoned score assignments~\citep{anugrahaR3RobustRubricAgnostic2025}.
Rubric-based methods provide an interpretable interface, but their reliability depends on rubric coverage and the judge model's interpretation of each criterion; rubrics can also be incomplete, redundant, overlapping, or misaligned with the intended preference direction~\citep{shenRethinkingRubricGeneration2026}.
This motivates complementary approaches that integrate multiple evaluative perspectives while grounding evaluation in human-preference supervision rather than judge-specific criterion interpretations.

\subsection{Multi-Expert Reward-Model Evaluation}
\label{sec:rw-multi-expert}

A growing line of work studies how to use multiple reward models rather than relying on a single evaluator.
Reward-model selection and routing methods choose among candidate reward models for each instance~\citep{nguyenLASeRLearningAdaptively2026,wuRewardModelRouting2025}, while scalar aggregation methods combine expert scores, such as reward-model ensembles~\citep{eisensteinHelpingHerdingReward2024} and log-sigmoid-centered reward aggregation~\citep{wangTransformingCombiningRewards2024}.
These methods show that heterogeneous reward models can provide complementary signals, but selection, routing, or scalar-score aggregation leaves limited room to model cross-expert interactions before scalar compression.

\subsection{Hidden Representations Beyond Final Outputs}
\label{sec:rw-hidden-representations}
Prior work has shown that neural models encode useful information in internal representations that may not be fully exposed by final outputs.
Analyses of Transformer representations show that different layers capture different linguistic and task-relevant properties~\citep{rogersPrimerBERTologyWhat2020}, and reward-model work has used hidden-state regularization to improve generalization under distribution shift~\citep{yangRegularizingHiddenStates2024}.
Related work on multi-model collaboration also suggests that internal expert representations can support prediction beyond final generated outputs or scalar decisions~\citep{fein-ashleyMixtureThoughtsLearning2025}.
However, how to align and fuse hidden states from multiple frozen reward experts for non-verifiable preference evaluation remains underexplored.

\section{Method}
\label{sec:method}

\begin{figure*}[t]
    \centering
    \includegraphics[width=\textwidth]{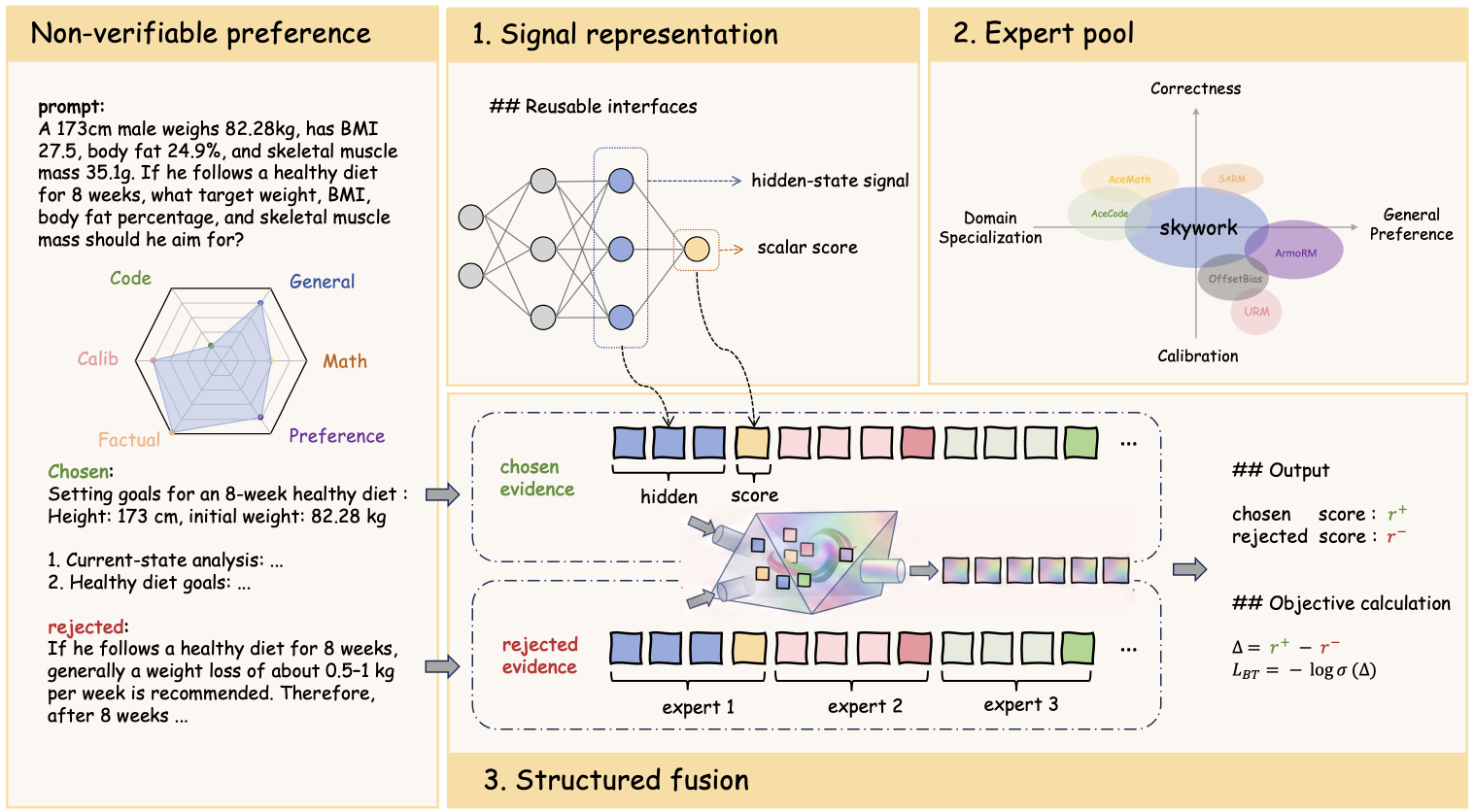}
    \caption{
    Overview of hidden-state evidence fusion from multiple reward experts for non-verifiable preference modeling.
    The figure abstracts three design factors in our study: signal interface, expert pool, and fusion structure.
    The structured fusion block is instantiated by \textsc{CSPF}.
    The radar chart and expert-pool map are illustrative.
    }
    \label{fig:overview}
\end{figure*}

To address the limitations of existing work, we propose Constrained Shared-Private Fusion (\textsc{CSPF}), a hidden-state fusion method that learns a target-domain reward function from frozen reward experts under pairwise human-preference supervision.
As shown in Figure~\ref{fig:overview}, the method is specified by three design factors: signal representation, expert pool, and fusion structure.
We detail these factors and the training objective below.

\subsection{Signal Representation}
\label{sec:expert-signals}

Let $x$ denote a prompt and $y$ a candidate response, and let
$E_k$, for $k \in \{1,\ldots,K\}$, denote the $k$-th of $K$
frozen reward experts.
Each expert exposes a scalar reward score $s_k(x,y) \in \mathbb{R}$.
Because raw score scales can differ across experts, we normalize
$s_k(x,y)$ using target-domain training statistics:
\begin{equation}
\tilde{s}_k(x,y)
=
\frac{s_k(x,y)-\mu_k}
{\max\{\sigma_k,\epsilon\}}
\label{eq:score-normalization}
\end{equation}
where $\mu_k$ and $\sigma_k$ are estimated on the training split
and fixed thereafter, and $\epsilon>0$ lower-bounds the denominator
for numerical stability.
The normalized score $\tilde{s}_k(x,y)$ is retained as an auxiliary
calibration signal.

Each expert also exposes a hidden representation before its final
scalar output.
Because these hidden states lie in model-specific representation
spaces and may differ in dimensionality, we map a selected
hidden-state readout $h_k(x,y) \in \mathbb{R}^{d_{k,h}}$ into a
common representation space:
\begin{equation}
\label{eq:summary-hidden-signal}
z_k(x,y)
=
P_k\!\left(\mathrm{LN}_k(h_k(x,y))\right)
\in \mathbb{R}^{d}
\end{equation}
where $\mathrm{LN}_k$ and $P_k$ are expert-specific normalization
and projection layers.
The projected representation $z_k(x,y)$ is the main signal consumed
by adapters and fusion modules.
Appendix~\ref{app:hidden-state-extraction} details the hidden-state
extraction protocol; Section~\ref{sec:signal-representation}
evaluates alternative layer and span choices.

\subsection{Expert Pool}
\label{sec:expert-pool}

We denote the candidate pool of $K$ pretrained reward experts by
$\mathcal{E}$ and the active expert index set for each configuration
by $C$:
\begin{equation}
\mathcal{E}
=
\{E_k\}_{k=1}^{K},
\qquad
C \subseteq \{1,\ldots,K\}
\label{eq:expert-pool}
\end{equation}
Only experts $E_k$ with $k \in C$ contribute the normalized score
$\tilde{s}_k(x,y)$ and projected hidden representation $z_k(x,y)$
defined in Section~\ref{sec:expert-signals} to fusion.
All expert backbones remain frozen; only modules built on these
signals are trained.

The set $C$ is fixed across examples rather than selected per
instance; sample-dependent interactions among expert signals are
modeled by the fusion module.
The experts are treated as complementary, potentially overlapping
evaluative perspectives, without assuming a one-to-one mapping to
human-defined criteria.
Concrete expert models and roles are described in
Section~\ref{sec:frozen-reward-expert-pool} and
Appendix~\ref{app:expert-pool}.

\subsection{Structured Fusion: \textsc{CSPF}}
\label{sec:cspf}

\begin{figure*}[t]
\centering
\includegraphics[width=0.98\textwidth]{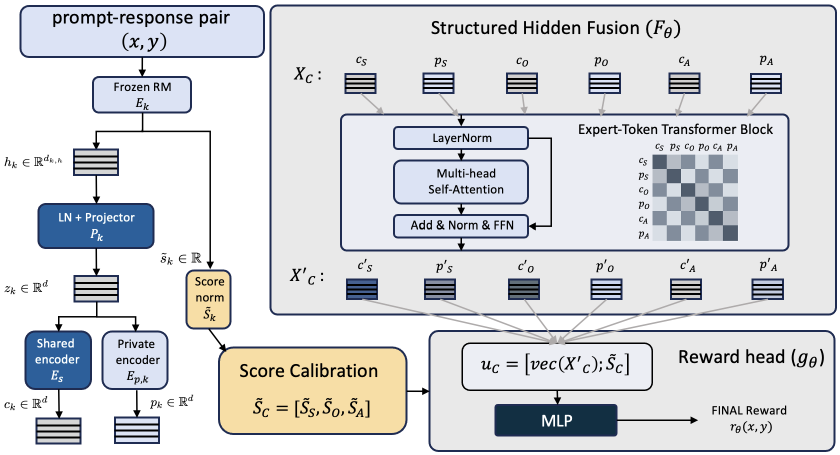}
\caption{
Architecture of \textsc{CSPF}.
The left branch shows the per-expert signal path; the right branch illustrates shared-private fusion over hidden representations from the active expert pool.
Normalized scalar scores are retained as auxiliary calibration signals.
}
\label{fig:cspf_architecture}
\end{figure*}

\textsc{CSPF} instantiates the fusion structure by decomposing each projected hidden representation into a cross-expert shared component and an expert-specific private component, as illustrated in Figure~\ref{fig:cspf_architecture}.

For each expert $E_k$, we apply a shared encoder $E_s$ and an expert-specific private encoder $E_{p,k}$:
\begin{equation}
\begin{array}{rcl}
c_k(x,y) & = & E_s\!\left(z_k(x,y)\right)\\[2pt]
p_k(x,y) & = & E_{p,k}\!\left(z_k(x,y)\right)
\end{array}
\label{eq:csp-factors}
\end{equation}
where $c_k$ captures shared preference-relevant information across experts and $p_k$ preserves expert-specific information.

For an active expert pool $C=\{k_1,\ldots,k_m\}$, we form a sequence of shared and private expert tokens,
\begin{equation}
X_C(x,y)
=
[c_{k_1},p_{k_1},\ldots,c_{k_m},p_{k_m}]
\label{eq:csp-tokens}
\end{equation}
and map it to a fused representation with a fusion module $\mathcal{F}_\theta$:
\begin{equation}
v_C(x,y)
=
\operatorname{vec}\!\left(\mathcal{F}_\theta(X_C(x,y))\right)
\label{eq:cspf-module}
\end{equation}
The final reward combines this fused hidden representation with normalized scalar scores used only as calibration signals:
\begin{equation}
\begin{aligned}
r_\theta(x,y)
&=
g_\theta\!\left([v_C(x,y);\tilde{s}_C(x,y)]\right)\\
\tilde{s}_C(x,y)
&=
[\tilde{s}_{k_1}(x,y);\ldots;\tilde{s}_{k_m}(x,y)] 
\end{aligned}
\label{eq:csp-reward}
\end{equation}
This reward function is trained with the pairwise preference objective and auxiliary constraints described next.

\subsection{Training Objective}
\label{sec:pairwise-adaptation}

We train all trainable evaluators on $N$ pairwise preference examples
$(x_i,y_i^+,y_i^-)$, where $y_i^+$ is preferred over $y_i^-$.
With margin
$\Delta_i=r_\theta(x_i,y_i^+)-r_\theta(x_i,y_i^-)$,
the primary task loss is the Bradley--Terry objective
\begin{equation}
\mathcal{L}_{\mathrm{BT}}
=
\frac{1}{N}
\sum_{i=1}^{N}
\operatorname{softplus}(-\Delta_i)
\label{eq:bt-loss}
\end{equation}

For \textsc{CSPF}, we further constrain the shared and private
representations.
For each response side $\rho\in\{+,-\}$, let
$\overline{\mathbf{C}}_k^\rho\in\mathbb{R}^{B\times d}$
collect the feature-wise standardized shared representations of
expert $E_k$ over a minibatch of $B$ preference pairs.
Let
$\mathcal{P}_C=\{(a,b):a,b\in C,\ a<b\}$.
We use a mean-reduced Barlow Twins-style objective
\citep{zbontarBarlowTwinsSelfSupervised2021}:
\begin{equation}
\begin{aligned}
R_{ab}^{\rho}
&=
B^{-1}
\left(\overline{\mathbf{C}}_{a}^{\rho}\right)^\top
\overline{\mathbf{C}}_{b}^{\rho},
\\
\mathcal{L}_{\mathrm{Barlow}}
&=
\mathbb{E}_{\rho,(a,b)}
\left[
\frac{
\left\|
\operatorname{diag}(R_{ab}^{\rho})-\mathbf{1}
\right\|_2^2
}{d}
\right.
\\[-0.2em]
&\hspace{5.2em}\left.
+
\frac{
\beta
\left\|
\operatorname{off}(R_{ab}^{\rho})
\right\|_F^2
}{d^2}
\right]
\end{aligned}
\label{eq:barlow-constraint}
\end{equation}
where $\operatorname{off}(\cdot)$ sets the diagonal to zero,
$\beta$ weights the off-diagonal penalty, and
$\mathbb{E}_{\rho,(a,b)}$ averages uniformly over the two response
sides and the expert pairs in $\mathcal{P}_C$.
The two sides are standardized separately before averaging.

For the private representations, let
$\mathcal{Q}=\{(q_u,e_u)\}_{u=1}^{M}$ collect the
$M=2B|C|$ representations from both response sides, with $e_u$
denoting source-expert identity.
With $P(u)=\{p\neq u:e_p=e_u\}$ and
$s_{ua}=\exp(\operatorname{sim}(q_u,q_a)/\tau)$, the supervised
contrastive objective~\citep{khoslaSupervisedContrastiveLearning2020}
is
\begin{equation}
\mathcal{L}_{\mathrm{SupCon}}
=
-\mathbb{E}_{u}
\mathbb{E}_{p\in P(u)}
\log
\frac{s_{up}}
{\sum_{a\neq u}s_{ua}}.
\label{eq:supcon-constraint}
\end{equation}
where both expectations are uniform,
$\operatorname{sim}(\cdot,\cdot)$ is cosine similarity, and $\tau$ is
the temperature.

With weights $\lambda_{\mathrm{B}}$ and $\lambda_{\mathrm{S}}$
for the shared- and private-space constraints, the complete
objective is
\begin{equation}
\mathcal{L}_{\mathrm{CSPF}}
=
\mathcal{L}_{\mathrm{BT}}
+
\lambda_{\mathrm{B}}\mathcal{L}_{\mathrm{Barlow}}
+
\lambda_{\mathrm{S}}\mathcal{L}_{\mathrm{SupCon}}
\label{eq:csp-objective}
\end{equation}
Thus, $\mathcal{L}_{\mathrm{BT}}$ provides task supervision to both
branches, while the two auxiliary losses impose shared- and
private-space structure, respectively.

\section{Experimental Setup}
\label{sec:experimental-setup}

\subsection{Datasets}
\paragraph{Target-domain adaptation and validation.}
We adapt and validate on a cleaned LM-Arena preference dataset derived
from crowdsourced Chatbot Arena comparisons of LLM
responses~\citep{chiangChatbotArenaOpen2024}, matching our
non-verifiable pairwise evaluation setting.

\paragraph{Out-of-distribution evaluation.}
We use the human-preference split of Preference Proxy Evaluations
(PPE)~\citep{frickHowEvaluateReward2025}, which targets
non-verifiable preferences rather than objective correctness.
PPE is excluded from training and score-normalization statistics and
used only for OOD evaluation.

\subsection{Frozen Reward-Expert Pool}
\label{sec:frozen-reward-expert-pool}
The expert pool contains seven frozen scalar reward models that expose both reward scores and hidden-state signals: Skywork-Reward, AceMath, OffsetBias, ArmoRM, AceCodeRM, SARM, and URM~\citep{liuSkyworkRewardV2ScalingPreference2025,liuAceMathAdvancingFrontier2025,parkOffsetBiasLeveragingDebiased2024,wangInterpretablePreferencesMultiObjective2024a,zengACECODERAcingCoder2025,zhangInterpretableRewardModel2026,louUncertaintyawareRewardModel2024}.
Skywork-Reward serves as the general anchor expert in single-expert baselines and controlled comparisons.
Each experiment uses an active subset of the candidate experts, denoted by $C$ in Section~\ref{sec:expert-pool}; the active subset is reported in the corresponding table or figure.
We focus on scalar reward models and exclude generative judges that require rationale or critique generation before scoring.
Detailed model versions and intended expert roles are summarized in Table~\ref{tab:appendix_expert_pool}.

\subsection{Baselines}
\label{sec:baselines}

\paragraph{Single-expert reward models.}
We evaluate individual reward-expert scalar scores directly.
Because \textsc{CSPF} is adapted on LM-Arena, we also adapt Skywork-Reward with LoRA on the same data, giving a matched parameter-update baseline with a comparable trainable-parameter budget: 15.34M for LoRA versus 17.45M for \textsc{CSPF}, excluding frozen backbones.
A frozen-backbone Skywork adapter is also included; it trains a
prediction head over the same hidden-state and normalized-score signal
interface as \textsc{CSPF}, but using a single expert.

\paragraph{Rubric evaluators.}
The rubric-family baselines include two open-source evaluators, Prometheus-2-7B and R3-Qwen3-8B-4k~\citep{kimPrometheus2Open2024,anugrahaR3RobustRubricAgnostic2025}.
Prometheus-2-7B represents a pairwise rubric-conditioned LLM judge, while R3-Qwen3-8B-4k represents a rubric-aware pointwise evaluator with scalar rubric scores.
Their pairwise and pointwise evaluation protocols are detailed in
Appendix~\ref{app:baseline-definitions}.

\paragraph{Multi-expert reward-model fusion.}
Using the same active expert pool as \textsc{CSPF}, RM Ensemble
averages normalized expert scores~\citep{eisensteinHelpingHerdingReward2024},
whereas LSC adapts log-sigmoid-centered aggregation to the
frozen-expert pairwise setting~\citep{wangTransformingCombiningRewards2024};
both fuse final scalar scores rather than hidden representations.
Definitions are provided in Appendix~\ref{app:baseline-definitions}.

\subsection{Metrics}
We use pairwise accuracy as the primary evaluation metric throughout.
On LM-Arena, we report validation accuracy as the target-domain metric.
On PPE, we report PPE off6, the equal-weighted average of pairwise accuracy over the six human-preference slices used in our evaluation, as the primary OOD metric.
This follows PPE's metric analysis, which identifies pairwise accuracy as a strong predictor of downstream post-RLHF human preference scores among human-preference metrics~\citep{frickHowEvaluateReward2025}.
Overall non-tie accuracy and slice-level PPE results are used as diagnostics when relevant.

\subsection{Training Details}
All trainable baselines and \textsc{CSPF} are optimized with the pairwise preference objective in Section~\ref{sec:pairwise-adaptation}.
For frozen-expert methods, we precompute scalar scores and selected hidden-state readouts on LM-Arena and PPE; the Skywork LoRA baseline is trained separately because it updates the reward-model backbone. Appendix~\ref{app:experimental-details} details hidden-state
extraction, training configurations, and baseline settings.

\section{Results and Analysis}
\label{sec:results}

\subsection{Main Results}
\label{sec:main-results}

\newcommand{\best}[1]{\(\bm{#1}\)}
\newcommand{\second}[1]{\(\underline{#1}\)}

\begin{table*}[t]
\centering
\scriptsize
\setlength{\tabcolsep}{2.55pt}
\renewcommand{\arraystretch}{1.16}
\resizebox{\textwidth}{!}{%
\begin{tabular}{@{}llcccccccccc@{}}
\toprule
\multirow{2}{*}{\makecell[c]{\textbf{Method}}}
& \multirow{2}{*}{\makecell[c]{\textbf{Interface}}}
& \multirow{2}{*}{\makecell[c]{\textbf{Experts}}}
& \multirow{2}{*}{\makecell[c]{\textbf{LM-Arena}\\\textbf{Val. Acc.}}}
& \multicolumn{8}{c}{\textbf{PPE OOD}} \\
\cmidrule(lr){5-12}
&
&
&
& \textbf{off6}
& \textbf{Hard}
& \textbf{Easy}
& \textbf{IF}
& \textbf{Code}
& \textbf{Math}
& \textbf{Sim.}
& \textbf{O/NT} \\
\midrule

\multicolumn{12}{@{}l}{\textit{Single-expert reward models}} \\
\addlinespace[0.15em]

OffsetBias-8B
& scalar score
& O
& 48.62
& 55.79
& 55.59
& 55.89
& 56.62
& 57.00
& 54.41
& 55.24
& 59.17 \\

ArmoRM-8B
& scalar score
& A
& 54.59
& 55.08
& 55.26
& 55.55
& 55.44
& 55.83
& 53.70
& 54.70
& 59.11 \\

Skywork-V2-8B
& scalar score
& S
& 60.43
& 60.24
& 60.37
& 61.29
& 60.38
& 60.60
& 59.03
& 59.78
& 67.59 \\

Skywork-V2 LoRA
& parameter update
& S
& 66.26
& 59.76
& 60.60
& 59.11
& 60.18
& 59.92
& \best{59.52}
& 59.27
& 67.02 \\

Skywork-V2 adapter (ours)
& hidden state
& S
& \second{66.84}
& \second{61.17}
& \second{61.60}
& \second{61.84}
& \second{61.09}
& \second{62.19}
& \second{59.35}
& \second{60.94}
& \second{68.78} \\

\midrule

\multicolumn{12}{@{}l}{\textit{Rubric evaluators}} \\
\addlinespace[0.15em]

Prometheus-2-7B
& natural language
& P
& 58.30
& 53.77
& 53.97
& 53.46
& 52.52
& 55.25
& 53.02
& 54.37
& 57.17 \\

R3-8B
& natural language
& R
& 56.22
& 53.49
& 53.92
& 52.48
& 54.15
& 53.57
& 54.19
& 52.61
& 54.69 \\

\midrule

\multicolumn{12}{@{}l}{\textit{Multi-expert reward-model fusion}} \\
\addlinespace[0.15em]

RM Ensemble
& scalar scores
& S+O+A
& 56.49
& 58.93
& 59.09
& 59.54
& 59.33
& 60.47
& 56.40
& 58.74
& 65.28 \\

LSC
& scalar scores
& S+O+A
& 56.58
& 59.54
& 60.00
& 59.94
& 59.76
& 61.50
& 56.72
& 59.35
& 66.13 \\

\textsc{CSPF} (ours)
& hidden states
& S+O+A
& \best{68.04}
& \best{61.67}
& \best{62.04}
& \best{62.33}
& \best{62.28}
& \best{63.11}
& 59.07
& \best{61.20}
& \best{69.17} \\

\bottomrule
\end{tabular}%
}
\caption{
Main comparison of \textsc{CSPF} with three families of evaluator baselines.
Values are pairwise accuracies (\%); \textsc{PPE} off6 averages the six PPE human-preference slices, and O/NT denotes overall non-tie accuracy.
S, O, A, P, and R denote Skywork, OffsetBias, ArmoRM, Prometheus, and R3, respectively.
Bold and underlined values mark the best and second-best results.
}
\label{tab:main_comparison}
\end{table*}

\begin{figure*}[t]
\centering
\includegraphics[width=0.88\textwidth]{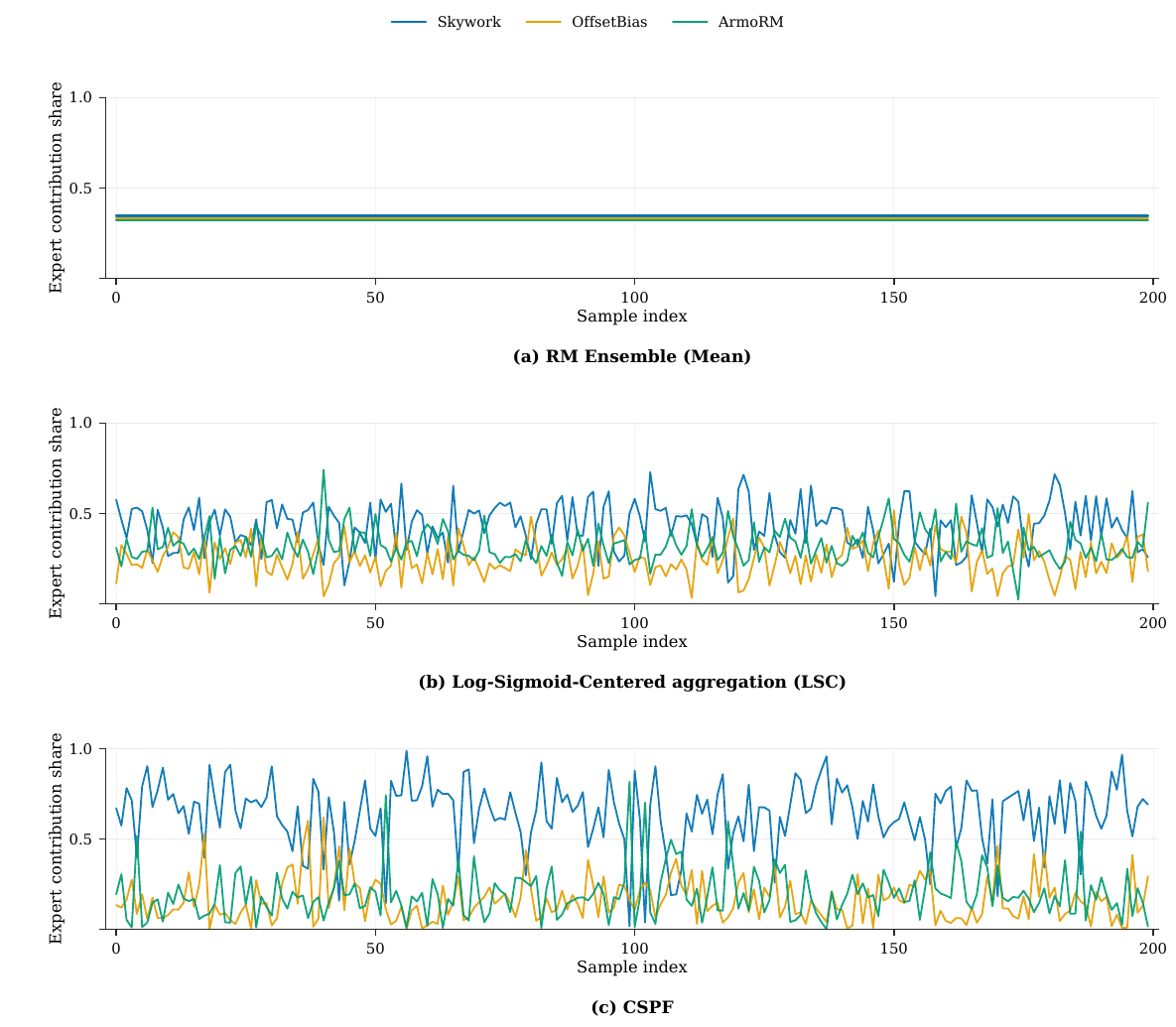}
\caption{
Sample-level expert-contribution shares on the same 200 random PPE
responses.
RM Ensemble is uniform by construction (coincident traces are slightly
offset for visibility); LSC and \textsc{CSPF} vary across responses,
with \textsc{CSPF} showing stronger sample-dependent shifts.
}
\label{fig:expert_use_shares}
\end{figure*}

Table~\ref{tab:main_comparison} reports the main comparison on
LM-Arena target-domain validation and PPE out-of-distribution
evaluation.
Overall, \textsc{CSPF} achieves the highest LM-Arena validation
accuracy (68.04) and PPE off6 accuracy (61.67) among all evaluated
methods, as well as the highest O/NT accuracy.

\textsc{CSPF} outperforms all raw single-expert reward models.
Skywork LoRA, trained on the same LM-Arena data, improves
target-domain accuracy but slightly lowers PPE off6 relative to raw
Skywork, revealing a target-domain/OOD trade-off rather than improved
OOD transfer.
Relative to Skywork LoRA, \textsc{CSPF} improves LM-Arena and PPE
off6 by 1.78 and 1.91 percentage points, respectively.
The frozen-backbone Skywork hidden+score adapter ranks second on both
primary metrics and O/NT, showing that augmenting a frozen expert with
hidden-state-based adaptation is effective even without multi-expert
fusion.
The additional gains of \textsc{CSPF} are consistent with a benefit
from integrating complementary expert signals.

For the two evaluated rubric models, \textsc{CSPF} also achieves
higher target-domain and OOD accuracy.
Their gap from the leading preference-trained reward models may
partly reflect a mismatch between a judge model's interpretation of
explicit criteria and the holistic judgments reflected in human
preferences.

RM Ensemble and LSC use the same active expert pool as \textsc{CSPF}
but fuse only final scalar scores; details are provided in
Appendix~\ref{app:baseline-definitions}.
\textsc{CSPF} outperforms LSC, the stronger multi-expert baseline,
by 11.46 and 2.13 percentage points on LM-Arena and PPE off6,
respectively.
Both multi-expert methods also remain below stronger single-expert
baselines, suggesting that scalar-score aggregation mainly smooths
or averages expert scores rather than producing complementary gains
beyond the strongest individual expert.

To further understand the fusion behavior behind these multi-expert methods,
we convert each method into a sample-level expert-contribution share
vector.
RM Ensemble has fixed uniform shares, LSC uses transformed scalar-score
contributions, and \textsc{CSPF} uses exact group-Shapley attribution
as a post-hoc diagnostic
\citep{jullumGroupShapleyEfficientPrediction2021}.
These shares are used only for diagnosis, not as learned router
weights; formal definitions are given in
Appendix~\ref{app:group-shapley}.

Figure~\ref{fig:expert_use_shares} visualizes sample-level
expert-contribution shares.
RM Ensemble is uniform by construction.
LSC shows moderate sample-level variation, but remains relatively
balanced across experts, behaving close to an averaging scheme.
By contrast, \textsc{CSPF} is overall more Skywork-dominant,
consistent with the strongest general reward model carrying greater
evaluative reliance.
At the same time, \textsc{CSPF} varies across samples: OffsetBias or
ArmoRM receives larger shares on some examples, indicating that
complementary experts can become more influential for particular
cases.
A complementary $L_1$-deviation analysis, reported in
Appendix~\ref{app:l1-deviation}, shows the same trend.

The above analysis suggests that \textsc{CSPF} does not merely
average frozen reward experts, but learns a more flexible
sample-dependent fusion of complementary expert signals.

\subsection{Component Ablation of \textsc{CSPF}}
\label{sec:cspf-component-ablation}

\begin{table*}[t]
\centering
\begingroup
\footnotesize
\setlength{\tabcolsep}{4pt}
\renewcommand{\arraystretch}{1.08}

\begin{tabular}{@{}lccccccc@{}}
\toprule
Configuration
& \shortstack{S/P\\enc.}
& \shortstack{Shared\\Barlow}
& \shortstack{Private\\SupCon}
& \shortstack{Score\\calib.}
& \shortstack{LM-Arena\\Val. Acc.}
& PPE off6
& O/NT \\
\midrule

\textsc{CSPF} w/o S/P encoders
& --
& --
& --
& $\checkmark$
& 67.95 $\pm$ 0.32
& 61.11 $\pm$ 0.66
& 68.73 $\pm$ 0.69 \\

\textsc{CSPF} w/o auxiliary constraints
& $\checkmark$
& --
& --
& $\checkmark$
& 68.03 $\pm$ 0.30
& 61.53 $\pm$ 0.24
& 69.03 $\pm$ 0.28 \\

\textsc{CSPF} w/o private SupCon
& $\checkmark$
& $\checkmark$
& --
& $\checkmark$
& 68.01 $\pm$ 0.26
& 61.62 $\pm$ 0.24
& 69.12 $\pm$ 0.20 \\

\textsc{CSPF} w/o score calibration
& $\checkmark$
& $\checkmark$
& $\checkmark$
& --
& 67.90 $\pm$ 0.17
& 61.37 $\pm$ 0.28
& 68.92 $\pm$ 0.59 \\

\textsc{CSPF}
& $\checkmark$
& $\checkmark$
& $\checkmark$
& $\checkmark$
& \textbf{68.04 $\pm$ 0.13}
& \textbf{61.67 $\pm$ 0.19}
& \textbf{69.17 $\pm$ 0.13} \\

\bottomrule
\end{tabular}

\endgroup
\caption{
Component ablation of \textsc{CSPF}.
All settings follow the main \textsc{CSPF} configuration except for the listed components.
Values are mean $\pm$ standard deviation over three seeds.
}
\label{tab:cspf_component_ablation}
\end{table*}

We ablate the main components of \textsc{CSPF}.
The \emph{Hidden fusion w/o S/P enc.} control uses the same frozen expert signals and scalar-score calibration path as \textsc{CSPF}, but feeds hidden representations directly into a fusion head.
We then cumulatively add shared/private encoders, the shared-space Barlow constraint, and the private SupCon constraint.

Table~\ref{tab:cspf_component_ablation} shows that shared/private encoders provide the primary OOD gain: replacing direct hidden fusion with shared/private factorization improves PPE off6 and O/NT, suggesting that organizing heterogeneous expert representations into shared and expert-specific factors is more transferable than direct concatenation.
The Barlow and SupCon constraints add smaller but consistent gains, while score calibration provides only modest improvement, indicating that normalized scalar scores mainly act as auxiliary calibration signals.

The ablation also shows improved seed-level stability.
Shared/private encoders reduce the standard deviation on PPE off6 and O/NT relative to direct hidden fusion, and both auxiliary constraints further reduce dispersion.
Overall, the shared/private fusion structure is the main factor behind \textsc{CSPF}'s gains, while auxiliary constraints and scalar-score calibration provide refinement.

\subsection{Signal Representation Analysis}
\label{sec:signal-representation}

\begin{table*}[t]
\centering
\begingroup
\footnotesize
\setlength{\tabcolsep}{7pt}
\renewcommand{\arraystretch}{1.08}

\begin{tabular}{p{0.36\textwidth}lccc}
\toprule
Hidden-state signal & Layer & LM-Arena Val. Acc. & PPE off6 & O/NT \\
\midrule

\multirow{7}{0.36\textwidth}{\textit{Last-nonpad}}
  & $0.25L$  & 66.52 & 55.26 & 61.05 \\
  & $0.375L$ & 67.09 & 56.84 & 62.89 \\
  & $0.5L$   & 67.93 & 58.56 & 65.41 \\
  & $0.625L$ & \textbf{68.28} & 59.83 & 66.72 \\
  & $0.75L$  & 67.84 & 61.04 & 68.58 \\
  & $0.875L$ & 67.77 & 60.58 & 67.86 \\
  & $L$      & 67.39 & 59.99 & 67.14 \\

\midrule

\multirow{7}{0.36\textwidth}{\textit{Last-nonpad + response mean}}
  & $0.25L$  & 67.07 & 55.73 & 61.62 \\
  & $0.375L$ & 67.01 & 57.15 & 63.50 \\
  & $0.5L$   & 67.93 & 58.52 & 65.48 \\
  & $0.625L$ & \underline{68.15} & 61.34 & 68.91 \\
  & $0.75L$  & 67.83 & 61.41 & \underline{69.10} \\
  & $0.875L$ & 67.81 & 61.47 & 68.94 \\
  & $L$      & 67.72 & 60.38 & 67.50 \\

\midrule

\multirow{7}{0.36\textwidth}{\textit{Last-nonpad + all-token mean}}
  & $0.25L$  & 67.24 & 55.71 & 61.78 \\
  & $0.375L$ & 67.09 & 57.11 & 63.38 \\
  & $0.5L$   & 67.47 & 58.38 & 65.33 \\
  & $0.625L$ & 67.89 & 61.23 & 68.85 \\
  & $0.75L$  & 67.97 & \underline{61.52} & \textbf{69.21} \\
  & $0.875L$ & 67.59 & \textbf{61.70} & 69.05 \\
  & $L$      & 67.21 & 60.95 & 68.02 \\

\bottomrule
\end{tabular}

\endgroup
\caption{
Signal representation analysis for \textsc{CSPF}.
All rows use S+O+A and vary only the hidden-state layer and span.
Intermediate layers are generally stronger, and the best layer--span choice differs across evaluation targets.
}
\label{tab:cspf_signal_interface}
\end{table*}

We analyze how hidden-state signal representation affects \textsc{CSPF}.
All settings follow the main \textsc{CSPF} configuration except for the hidden-state readout. Table~\ref{tab:cspf_signal_interface} varies only the hidden-state
readout depth and span; $L$ denotes the final block, and fractional
values denote depth-normalized intermediate blocks.

Table~\ref{tab:cspf_signal_interface} shows two patterns.
First, final-layer signals are not strongest: LM-Arena peaks at the intermediate $0.625L$ last-nonpad signal, while very early signals transfer poorly to PPE.
This suggests that useful expert information is concentrated in intermediate layers, after preference-relevant abstraction forms but before final scalar-reward compression.
Second, the preferred representation differs by evaluation target: LM-Arena favors a compact mid-layer signal, whereas PPE benefits more from pooled span-level evidence at later intermediate layers.
Thus, effective fusion depends not only on which experts are used, but also on which internal representations are exposed.

\subsection{Expert-Pool Composition Analysis}
\label{sec:expert-pool-composition}

\begin{figure*}[t]
\centering
\includegraphics[width=0.92\textwidth]{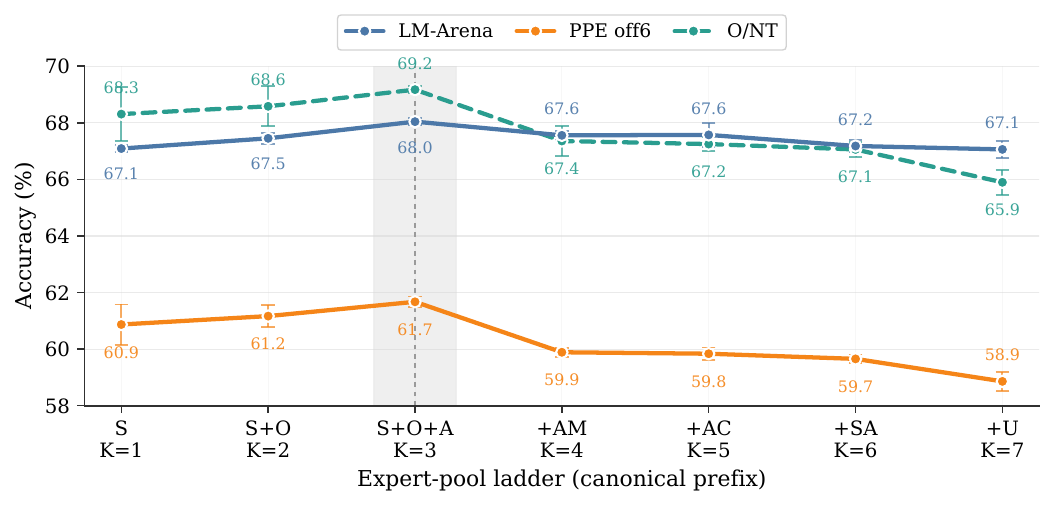}
\caption{
Expert-pool ladder ablation for \textsc{CSPF}.
+AM, +AC, +SA, and +U cumulatively add AceMath, AceCodeRM, SARM, and URM after S+O+A.
Points and error bars show three-seed means and standard deviations.
}
\label{fig:cspf_expert_pool_ladder}
\end{figure*}

We examine expert-pool composition by varying only the active pool
under a fixed \textsc{CSPF} structure.
Starting from the Skywork anchor, we add experts along the cumulative
ladder in Figure~\ref{fig:cspf_expert_pool_ladder} to test whether
pool expansion yields monotonic gains.

Performance improves from S to S+O and peaks at S+O+A, the pool used
by the main \textsc{CSPF} configuration.
All subsequent larger pools score below S+O+A on all three metrics,
with the largest drops on PPE off6 and O/NT.
Thus, adding experts can improve performance, but expert count alone
is not a reliable scaling rule.
Within \textsc{CSPF}, expert-pool expansion is therefore a
system-level design decision that should account for both the fusion
structure and intended application domain.

\section{Discussion}
\label{sec:discussion}

\paragraph{Implicit Multi-Perspective Evaluation.}
\textsc{CSPF} can be viewed as an implicit multi-perspective evaluator for non-verifiable preference tasks.
Such tasks often depend on multiple coupled criteria that are difficult to specify, weight, or aggregate explicitly.
Rather than asking an LLM judge to interpret natural-language rubrics, \textsc{CSPF} learns under pairwise human-preference supervision to fuse hidden representations from multiple frozen reward experts.
This does not imply that \textsc{CSPF} recovers explicit human criteria; instead, it provides a practical way to integrate latent evaluative factors that may underlie composite human preferences.

Taken together, the results position \textsc{CSPF} between three evaluator families.
Like holistic single-expert reward models, \textsc{CSPF} is trained toward overall human preferences through pairwise supervision; unlike them, it integrates multiple frozen evaluators.
Like rubric-based judges, it reflects the need for multiple evaluative perspectives, but obtains these perspectives from reward-expert representations rather than explicit natural-language criteria.
Compared with scalar-score multi-expert fusion, it models interactions among latent evaluative factors at the representation level.
This combination helps explain why \textsc{CSPF} achieves stronger non-verifiable preference evaluation on both LM-Arena and PPE.

\paragraph{Multi-Expert Fusion as a Coupled Design.}
Our analyses show that, in the \textsc{CSPF} framework, multi-expert hidden-state fusion is a coupled design problem.
The signal representation determines which expert information is exposed, the expert pool determines which evaluative perspectives participate, and the fusion structure determines how heterogeneous representations are organized.
Thus, hidden representations provide a richer interface than final scalar rewards, but effective fusion still requires coordinated choices about where representations are extracted, which experts are included, and how their signals are fused.

\paragraph{Limitations.}
Several limitations remain.
This work studies \textsc{CSPF} as a static pairwise evaluator, using LM-Arena for adaptation and PPE for OOD evaluation; whether its gains transfer to downstream settings, including best-of-$N$ selection, DPO, and other RLHF-style policy-optimization pipelines, remains to be tested.
\textsc{CSPF} also focuses on scalar reward models that expose hidden states, leaving generative judges and critique-then-score evaluators outside the main fusion setting.
Future work should investigate these directions to better understand and extend multi-expert hidden-state fusion.

\section{Conclusion}
\label{sec:conclusion}

This work examined how to build evaluators for non-verifiable preference tasks, where open-ended LLM responses must be judged through composite human preferences rather than deterministic correctness signals.
Motivated by this direction, we proposed \textsc{CSPF}, a non-verifiable preference evaluation method that fuses multiple frozen reward experts.
The method integrates complementary expert perspectives at the hidden-representation level and implicitly models interactions among expert signals through constrained shared-private fusion.
Empirically, \textsc{CSPF} yields the strongest primary results among the evaluated evaluator families.
The analyses further show that the gains depend not only on using multiple experts, but also on coordinated choices of signal representation, expert-pool composition, and fusion structure.
Overall, our results suggest that hidden-state fusion is a practical representation-level interface for constructing integrated evaluative signals for non-verifiable preference evaluation.

\bibliographystyle{acl_natbib}
\bibliography{references}

\appendix
\section*{Appendix}

\makeatletter
\@addtoreset{equation}{section}
\@addtoreset{table}{section}
\@addtoreset{figure}{section}
\makeatother

\renewcommand{\theequation}{\thesection.\arabic{equation}}
\renewcommand{\thetable}{\thesection.\arabic{table}}
\renewcommand{\thefigure}{\thesection.\arabic{figure}}

\section{Experimental details}
\label{app:experimental-details}

\subsection{Frozen reward expert pool}
\label{app:expert-pool}

Table~\ref{tab:appendix_expert_pool} summarizes the frozen reward experts considered in our study.

\begin{table*}[t]
\centering
\small
\setlength{\tabcolsep}{4pt}
\renewcommand{\arraystretch}{1.05}
\begin{tabularx}{\textwidth}{
@{}
l
c
>{\raggedright\arraybackslash}X
>{\raggedright\arraybackslash}p{0.23\textwidth}
@{}
}
\toprule
\textbf{Expert}
& \textbf{Abbrev.}
& \textbf{Model version}
& \textbf{Role} \\
\midrule

Skywork
& S
& Skywork-Reward-V2-Qwen3-8B~\citep{liuSkyworkRewardV2ScalingPreference2025}
& general-purpose anchor \\

AceMath
& AM
& AceMath-7B-RM~\citep{liuAceMathAdvancingFrontier2025}
& math-oriented \\

OffsetBias
& O
& Llama-3-OffsetBias-RM-8B~\citep{parkOffsetBiasLeveragingDebiased2024}
& bias-aware preference \\

ArmoRM
& A
& ArmoRM-Llama3-8B-v0.1~\citep{wangInterpretablePreferencesMultiObjective2024a}
& multi-objective preference \\

AceCodeRM
& AC
& AceCodeRM-7B~\citep{zengACECODERAcingCoder2025}
& code-domain \\

SARM
& SA
& Llama-SARM-4B~\citep{zhangInterpretableRewardModel2026}
& factuality / focus \\

URM
& U
& URM-LLaMa-3.1-8B~\citep{louUncertaintyawareRewardModel2024}
& attribute / calibration \\

\bottomrule
\end{tabularx}
\caption{
Frozen scalar reward experts considered in our study.
Depending on the experiment, we use their reward scores, hidden states, or both.
}
\label{tab:appendix_expert_pool}
\end{table*}

\subsection{Hidden-state signal extraction}
\label{app:hidden-state-extraction}

For token-level hidden-state signals, \emph{last-nonpad} denotes the final non-padding token representation, \emph{response mean} denotes mean pooling over assistant-response tokens, and \emph{all-token mean} denotes mean pooling over non-padding prompt-response tokens.
When multiple hidden-state signals are used together, they are concatenated before layer normalization and expert-specific projection.
Layer positions are specified as proportions of the expert backbone depth; for example, $0.75L$ denotes the proportional layer at three quarters of the transformer block stack.

\subsection{Training hyperparameters}
\label{app:training-hyperparameters}

Table~\ref{tab:training_hyperparameters} summarizes the default training configuration used for the main \textsc{CSPF} row and matched ablations.
Unless otherwise specified, all trainable adapters and fusion heads are trained on LM-Arena with frozen reward experts and evaluated with the same LM-Arena validation and PPE protocols as in Section~\ref{sec:experimental-setup}.
Score-normalization statistics are estimated only on the LM-Arena training split and then fixed for validation and OOD evaluation.

\begin{table}[t]
\centering
\small
\setlength{\tabcolsep}{4pt}
\renewcommand{\arraystretch}{1.05}
\begin{tabular}{@{}p{0.40\columnwidth}p{0.52\columnwidth}@{}}
\toprule
\textbf{Item} & \textbf{Value} \\
\midrule
Active expert pool & S+O+A \\
Hidden-state signal & last-nonpad + response mean \\
Layer specification & $0.75L$ proportional layer \\
Fusion structure & shared/private encoders + structured hidden fusion \\
Score calibration & final concat of normalized scalar scores \\
Main objective & Bradley--Terry pairwise loss \\
Epochs & 3 \\
Batch size & 128 \\
Learning rate & $1\times10^{-4}$ \\
Weight decay & $1\times10^{-2}$ \\
Dropout & 0.1 \\
Gradient clipping & 1.0 \\
Seeds & 40, 43, 45 \\
\makecell[l]{Shared-space\\constraint}
& Barlow loss ($\lambda_{\mathrm{B}}=1\times10^{-3}$,
  $\beta=0.005$) \\
\makecell[l]{Private-space\\constraint}
& SupCon loss ($\lambda_{\mathrm{S}}=1\times10^{-4}$,
  $\tau=0.10$) \\
Score normalization & LM-Arena train split only \\
Reward experts & frozen \\
\bottomrule
\end{tabular}
\caption{
Default training configuration for the main \textsc{CSPF} row and matched ablations.
}
\label{tab:training_hyperparameters}
\end{table}

\subsection{Baseline definitions}
\label{app:baseline-definitions}

This appendix defines the baselines used in the main comparison.
Frozen-expert baselines use one or both of the scalar-score and
hidden-state interfaces defined in
Section~\ref{sec:expert-signals}, and all trainable baselines are
optimized with the pairwise objective in
Section~\ref{sec:pairwise-adaptation}.

\paragraph{Raw scalar reward.}
The raw-score baseline directly uses the scalar output of a single reward expert:
\begin{equation}
r_{\mathrm{raw},k}(x,y)
=
s_k(x,y)
\label{eq:app-raw-score}
\end{equation}

\paragraph{Single-expert adapter.}
For frozen single-expert adaptation, the reward expert remains fixed and we train a prediction head on top of its hidden-state and normalized scalar-score signals:
\begin{equation}
r_{\mathrm{adapter},k}(x,y)
=
f_\theta\!\left([z_k(x,y);\tilde{s}_k(x,y)]\right)
\label{eq:app-single-hidden-adapter}
\end{equation}
The adapter uses the optimizer settings in
Table~\ref{tab:training_hyperparameters} but is trained for one epoch;
the reported result is averaged over seeds 40--45.
For the LoRA baseline, we instead adapt the $q/k/v/o$ attention
projections ($r=16$, $\alpha=32$, dropout $=0.05$) together with the
native reward head of Skywork.
It is trained with the same pairwise objective for one epoch using a
learning rate of $2\times10^{-5}$; the reported result uses seed 43.

\paragraph{RM Ensemble (Mean).}
RM Ensemble (Mean) adapts standard reward-model ensemble aggregation to our frozen heterogeneous-expert setting by averaging normalized scalar scores over the active expert pool:
\begin{equation}
r_{\mathrm{mean}}(x,y)
=
\frac{1}{|C|}
\sum_{k\in C}
\tilde{s}_k(x,y)
\label{eq:app-rm-mean}
\end{equation}

\paragraph{LSC aggregation.}
LSC adapts log-sigmoid-centered reward aggregation to the same frozen expert pool.
For each expert, let $q_k$ be the median normalized score over LM-Arena training responses:
\begin{equation}
q_k
=
\operatorname{median}_{(x,y)\in \mathcal{D}_{\mathrm{train}}}
\tilde{s}_k(x,y)
\label{eq:app-lsc-median}
\end{equation}
The LSC reward is
\begin{equation}
r_{\mathrm{LSC}}(x,y)
=
\sum_{k\in C}
\log \sigma\!\left(\tilde{s}_k(x,y)-q_k\right)
\label{eq:app-lsc}
\end{equation}

\paragraph{Rubric-evaluator protocols.}
Both rubric baselines use fixed, dataset-independent general-quality
rubrics.
For each baseline, the same rubric is applied to LM-Arena and PPE
without dataset- or slice-specific modification.
Figure~\ref{fig:rubric-evaluator-prompts} presents the exact rubric
text and required decision/score formats used in our experiments.

\begin{figure*}[t]
\centering

\begin{rubricbox}{Prometheus-2: Pairwise Evaluation}
\textbf{Reference-answer field}

No reference answer is provided because this is a non-verifiable
preference task.
Judge the responses only according to the instruction and the score
rubric.

\medskip
\textbf{Evaluation rubric}

Choose the response that a human evaluator would prefer for the given
instruction.
Consider: (1) instruction following and responsiveness to the user's
request; (2) correctness, factuality, and sound reasoning;
(3) completeness, usefulness, and actionable detail;
(4) safety, appropriateness, and avoidance of harmful or misleading
advice; and (5) clarity, organization, and style.
For math, code, or technical tasks, give special weight to correctness
and executable reasoning.
When criteria conflict, prefer the response that is more helpful and
reliable overall.

\medskip
\textbf{Output constraint}

Return the final decision as \texttt{[RESULT] A} or
\texttt{[RESULT] B}.
\end{rubricbox}

\vspace{0.8em}

\begin{rubricbox}{R3: Pointwise Evaluation}
\textbf{Evaluation rubric}

Use the following 1--5 scale to judge how good the response is for the
instruction.

\textbf{5 = Excellent:} fully follows the instruction, is correct and
reliable, complete, useful, safe, and clearly written.

\textbf{4 = Good:} mostly follows the instruction and is useful, with
only minor omissions or weaknesses.

\textbf{3 = Fair:} partially useful but has noticeable omissions, weak
reasoning, unclear presentation, or limited helpfulness.

\textbf{2 = Poor:} substantially incomplete, unhelpful, unsafe, or
contains significant mistakes.

\textbf{1 = Very poor:} fails the task, is mostly incorrect, harmful,
or irrelevant.

Consider instruction following, correctness/factuality, reasoning
quality, completeness, usefulness, safety, and clarity.
For math, code, or technical tasks, give special weight to correctness
and executable reasoning.
For open-ended non-verifiable tasks, prefer the response that would be
more helpful and reliable to a human user.

\medskip
\textbf{Output constraint}

Begin the output with
\texttt{Score: <one number from 1 to 5>}, followed by a brief
assessment.
\end{rubricbox}

\caption{
Exact rubric text and required decision formats used for the two
rubric evaluators.
Prometheus-2 compares a response pair directly, whereas R3 scores the
two responses independently.
}
\label{fig:rubric-evaluator-prompts}
\end{figure*}

For Prometheus-2, response position is balanced by deterministic
hashing, and unparsed decisions receive half credit.
A reverse-order audit on 10\% of the examples is used only for
diagnosis.
For R3, the higher-scoring response is predicted as preferred; equal
scores and unparsed outputs receive half credit.
Prometheus-2 uses greedy decoding, whereas R3 uses non-thinking
decoding with temperature 0.6, top-$p$ 0.95, and top-$k$ 20; both
use at most 512 generated tokens, with input budgets of 7,168 and
8,192 tokens, respectively.

\section{Expert-Use Share Diagnostics}
\label{app:group-shapley}

This appendix defines post-hoc expert-contribution shares for RM
Ensemble, LSC, and \textsc{CSPF}; these diagnostics are neither
learned router weights nor part of training or inference.

\subsection{Scalar-Score Contribution Shares}

For a sample $(x_i,y_i)$, RM Ensemble assigns each active expert a
fixed uniform share:
\begin{equation}
u_{i,k}^{(\mathrm{Mean})}
=
\frac{1}{|C|},
\qquad k\in C
\label{eq:mean-expert-share}
\end{equation}

For LSC, let
\begin{equation}
\ell_{i,k}
=
\log\sigma\!\left(\tilde{s}_k(x_i,y_i)-q_k\right)
\label{eq:lsc-expert-contribution}
\end{equation}
where $q_k$ is the training-set reference defined in
Appendix~\ref{app:baseline-definitions}.
We use the normalized absolute transformed-score contribution
\begin{equation}
u_{i,k}^{(\mathrm{LSC})}
=
\frac{|\ell_{i,k}|}
{\sum_{j\in C}|\ell_{i,j}|}
\label{eq:lsc-expert-share}
\end{equation}

\subsection{\textsc{CSPF} Group-Shapley Shares}
\label{app:shapley-values}

For \textsc{CSPF}, the shared representation, private representation,
and normalized scalar score from each active expert form one group.
For each sample, we compute exact Group-Shapley values over these
groups~\citep{jullumGroupShapleyEfficientPrediction2021}, using
$F_i(T)=r_\theta^{(T)}(x_i,y_i)$ as the coalition value and replacing
groups outside $T$ with their LM-Arena training-split means.
For S+O+A, all eight coalitions are evaluated.

We convert the signed values to normalized absolute contribution
shares
\begin{equation}
p_{i,k}
=
\frac{|\phi_{i,k}|}
{\sum_{j\in C}|\phi_{i,j}|}
\label{eq:app-shapley-share}
\end{equation}
After excluding zero-denominator samples, we set
$\mathbf{u}_i^{(\mathrm{CSPF})}=(p_{i,k})_{k\in C}$.

\subsection{\(L_1\) Deviation from Global Expert-Use Mean}
\label{app:l1-deviation}

To quantify sample-level variation in expert use, we compute the
\(L_1\) distance between each sample's expert-use share vector and
the method-specific global mean.
For a method $m$ with $N$ valid samples, let
$\mathbf{u}_i^{(m)}\in\mathbb{R}^{|C|}$ denote its normalized
expert-use share vector and let
\begin{equation}
\bar{\mathbf{u}}^{(m)}
=
\frac{1}{N}
\sum_{i=1}^{N}
\mathbf{u}_i^{(m)}
\label{eq:app-global-share-mean}
\end{equation}

The sample-level deviation is
\begin{equation}
d_i^{(m)}
=
\left\|
\mathbf{u}_i^{(m)}
-
\bar{\mathbf{u}}^{(m)}
\right\|_1
\label{eq:app-l1-deviation}
\end{equation}

\begin{figure}[t]
\centering
\includegraphics[width=0.95\columnwidth]{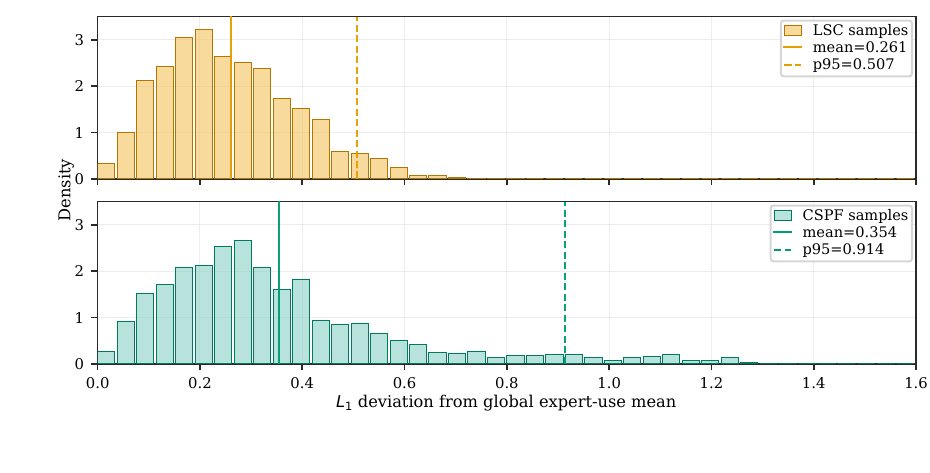}
\caption{
Distribution of sample-level expert-use variation on 2000 randomly
sampled PPE responses.
The \(L_1\) distance is computed between each sample's expert-use
share vector and the method-specific global mean.
\textsc{CSPF} has a more dispersed and longer-tailed distribution
than LSC.
}
\label{fig:app_l1_deviation}
\end{figure}

A larger value indicates stronger deviation from the method's
average expert-use pattern.
RM Ensemble has zero deviation by construction and is therefore
omitted from Figure~\ref{fig:app_l1_deviation}.

\end{document}